\documentclass{article}

\usepackage{arxiv}

\usepackage[utf8]{inputenc} % allow utf-8 input
\usepackage[T1]{fontenc}    % use 8-bit T1 fonts
\usepackage{hyperref}       % hyperlinks
\usepackage{url}            % simple URL typesetting
\usepackage{booktabs}       % professional-quality tables
\usepackage{amsfonts}       % blackboard math symbols
\usepackage{nicefrac}       % compact symbols for 1/2, etc.
\usepackage{microtype}      % microtypography
\usepackage{lipsum}		% Can be removed after putting your text content
\usepackage{graphicx}
\usepackage{natbib}
\usepackage{doi}
\usepackage{amsmath}
\usepackage{amsthm}
\usepackage{multirow}

\theoremstyle{definition}
\newtheorem{definition}{Definition}

\title{Modular Foundation Models for Time-Series Perception in Digital Twins}

\author{ {Quang Hung Pham}\thanks{Corresponding author : Pham.QuangHung2@hydroquebec.com} \\
	Hydro-Québec\\
    75, boulevard René-Lévesque Ouest\\
    Montréal, Québec, Canada\\
	\And
	{Ryad Zemouri} \\
	Hydro-Québec\\
    75, boulevard René-Lévesque Ouest\\
    Montréal, Québec, Canada\\
    \And
	{Martin Gagnon} \\
	Hydro-Québec\\
    75, boulevard René-Lévesque Ouest\\
    Montréal, Québec, Canada\\
    \And
	{Luc Vouligny} \\
	Hydro-Québec\\
    75, boulevard René-Lévesque Ouest\\
    Montréal, Québec, Canada\\
}

\hypersetup{
pdftitle={Modular Foundation Models for Time-Series Perception in Digital Twins: Toward Hybrid PHM Systems},
pdfsubject={q-bio.NC, q-bio.QM},
pdfauthor={Quang Hung Pham, Ryad Zemouri, Martin Gagnon, Luc Vouligny},
}

\begin{document}
\maketitle

\begin{abstract}
	Engineering Digital Twins and PHM systems require robust perception modules capable of extracting actionable information from heterogeneous, non-stationary time-series data, yet existing methods are often task-specific, data-hungry, and difficult to transfer across operating conditions. To address this, this paper proposes a modular foundation model based on a collection of pretrained representation encoders learned through self-supervised learning on heterogeneous datasets, enabling transferable and task-agnostic representations for multiple PHM tasks. A gating mechanism dynamically selects the most relevant encoders for each target dataset, while a Transformer-based self-attention module aligns and aggregates the selected representations in a shared latent space. The framework supports diverse downstream tasks, including imputation, long-term forecasting, and few-shot learning, through lightweight task-specific heads. Experiments on the ETT benchmark and a real-world virtual sensing case for hydro-generator rotor temperature demonstrate competitive performance and highlight the potential of the proposed approach as a scalable perception layer for industrial digital twin and hybrid model-data PHM applications.
\end{abstract}

% keywords can be removed

\section{Introduction}\label{sec1}
Digital Twins (DTs) are increasingly adopted across industrial domains as computational counterparts of physical assets that evolve in synchrony with real-world processes. 
By continuously mirroring the state and behavior of physical systems, DTs enable monitoring, forecasting, diagnostics, and informed decision-making throughout the asset lifecycle. 
As in any cyber--physical system, the most critical stage in the development of a Digital Twin is data perception, which provides the foundation for subsequent modeling, learning, and reasoning processes~\cite{fett2025survey}.

In industrial environments, DT perception relies on multivariate time-series data from heterogeneous sensing infrastructures, which face multiple interconnected challenges: data streams often differ in sampling rates, signal quality, and reliability; labeling is limited and data suffers from class imbalance; and systems operate under non-stationary regimes influenced by load fluctuations, component aging, environmental disturbances, and evolving operational conditions. These constraints make traditional supervised and task-specific learning pipelines costly to develop and difficult to maintain. This problem intensifies when building Engineering Digital Twins (EDTs) for safety-critical and performance-sensitive systems such as industrial plants, energy systems, communication networks, and smart infrastructures, where these challenges are compounded by stringent requirements on robustness, scalability, and interpretability.

Mixture-of-Experts (MoE) architectures provide a principled framework for addressing several of these challenges~\cite{Cai2025}. 
MoE models consist of multiple expert networks whose outputs are selectively combined through a gating mechanism, enabling conditional computation, modularity, and specialization. 
Originally introduced in the context of neural network learning~\cite{jacobs1991adaptive}, MoE architectures have demonstrated strong scalability and adaptability in large-scale and structured learning tasks~\cite{vats2024survey}. 
Originally introduced in the context of neural network learning~\cite{jacobs1991adaptive}, MoE architectures have demonstrated strong scalability and adaptability in large-scale and structured learning tasks~\cite{vats2024survey}. 
These properties make them particularly well suited for Digital Twin applications, where individual experts can naturally represent distinct operating regimes, physical subsystems, or functional behaviors.

Industrial engineering systems often exhibit pronounced regime-dependent dynamics driven by operating conditions, degradation mechanisms, and control strategies. 
To capture this variability, Zhou \emph{et al.} proposed MoE-based global modeling approaches in which experts specialize over different operating ranges, resulting in improved robustness and generalization for predictive maintenance and process optimization tasks~\cite{Zhou2025}. 
More generally, recent reviews on industrial Digital Twins highlight the importance of modular learning architectures for modeling multi-scale physical phenomena and enabling extensible system designs~\cite{nele2024review}. 
Emerging research further positions MoE as a unifying abstraction for multi-domain Digital Twins, where distinct experts may model physical dynamics, sensing uncertainty, cyber delays, or control responses. 
Such modularity supports incremental development and real-time adaptation, which are critical requirements for large-scale engineering deployments~\cite{fett2025survey}.

A recent study in~\cite{JOSE2026132252, Zemouri2025b} proposed an end-to-end modular deep learning methodology to address data quality, uncertainty, and scalability challenges in industrial diagnostics. 
By integrating tailored MoE networks with dynamic gating mechanisms, the approach demonstrated improved adaptability and computational efficiency for multimodal fault detection on a real-world hydrogenerator fleet.

Motivated by the data perception challenges inherent to Engineering Digital Twins and the limitations of existing task-specific learning pipelines, this paper proposes a modular foundation model for time-series perception built upon a collection of pretrained representation encoders. Unlike conventional end-to-end architectures, the proposed framework explicitly decouples representation learning, encoder selection, and task adaptation. Each encoder is pretrained in a self-supervised manner on heterogeneous time-series datasets to extract transferable and task-agnostic representations. A gating mechanism is then introduced to dynamically select a subset of relevant encoders for a given target dataset, enabling conditional computation at the encoder level. The selected representations are projected into a shared latent space and aggregated using a Transformer-based self-attention module that models cross-encoder interactions. This design allows a single foundation model to support multiple downstream tasks—such as imputation, forecasting, and virtual sensing—through lightweight task-specific heads, while keeping pretrained encoders frozen during adaptation. The proposed framework can also be interpreted as a perception layer within hybrid PHM architectures, where data-driven representations are combined with physics-based models to enhance robustness, interpretability, and scalability in practical industrial deployments.

The remainder of this paper is organized as follows. Section~II reviews Mixture-of-Experts models for time-series forecasting and situates the proposed approach within the broader literature. Section~III introduces the notation and formal definitions used throughout the paper. Section~IV presents the proposed foundation model architecture, including encoder pretraining, gating and selection, representation projection, and adaptive aggregation. Section~V reports extensive ablation studies analyzing the contribution of each architectural component. Section~VI presents benchmark evaluations and a real-world industrial case study on virtual sensing for hydro-generator rotor temperature. Finally, Section~VII concludes the paper and outlines directions for future research.

\section{Mixture-of-Experts Models for Time Series Forecasting}\label{sec2}
Accurate time-series forecasting remains a central challenge in machine learning, particularly in the presence of non-stationarity, multi-scale temporal dependencies, and regime shifts. 
In industrial monitoring contexts, two empirical observations consistently arise. 
First, large volumes of unlabeled operational data are routinely available. 
Second, many local temporal regularities, such as short-range dynamics and recurring patterns, are shared across assets, sites, and operating conditions. 
These observations motivate learning paradigms in which general-purpose representations are learned in an unsupervised manner and subsequently adapted using limited labeled data for asset-specific or task-specific objectives.

This paradigm aligns with the pretrain--adapt strategy enabled by self-supervised learning (SSL), which has gained significant momentum in large-scale time-series modeling~\cite{zhang_self-supervised_2024,das_decoder-only_2024,SelfSupervised_Jing2021}. 
In contrast, conventional forecasting models often rely on a single global predictor, which limits their capacity to capture heterogeneous and regime-dependent dynamics.

MoE models address this limitation by combining multiple specialized predictors through an adaptive gating mechanism. 
Early applications of MoE to time-series prediction focused on mixtures of linear autoregressive models. 
A seminal contribution by Zeevi \emph{et al.}~\cite{zeevi1998time} established that mixtures of autoregressive experts act as universal approximators for nonlinear time-series predictors and provided theoretical bounds on approximation and prediction errors.

Recent advances in deep learning have renewed interest in MoE architectures for time-series forecasting. 
In modern formulations, experts are implemented as neural networks, including recurrent, convolutional, or linear architectures, and are trained jointly with the gating network. 
A notable example is the mixture-of-linear-experts framework in~\cite{ni2024mixture}, which shows that even simple linear experts can achieve state-of-the-art long-term forecasting performance when combined with adaptive routing mechanisms. 
This result highlights the importance of expert diversity and gating dynamics over individual model complexity.

Gu \emph{et al.}~\cite{Gu2026} proposed a noise-conditioned MoE framework for robust speaker verification, where routing decisions enhance robustness across diverse noise conditions. 
The gating mechanism automatically directs inputs to specialized experts using noise-related cues, enabling each expert to model specific acoustic characteristics while preserving speaker-discriminative information. 
Similarly, Liang \emph{et al.}~\cite{Liang2026} introduced a multi-gate MoE framework with heterogeneous neural architectures to jointly capture local and global temporal features in non-intrusive load monitoring.

Transformer-based architectures have further enabled MoE models to scale to unprecedented sizes. 
Although not explicitly framed as a MoE, the Temporal Fusion Transformer~\cite{lim2021temporal} introduced gating and variable selection mechanisms that inspired subsequent work on conditional computation in time-series forecasting. 
The GShard architecture~\cite{lepikhin2020} generalized sparse MoE as a systems-level solution for conditional computation and automatic sharding, enabling large-scale distributed training with limited emphasis on expert behavior. 
Switch Transformers~\cite{fedus2022switchtransformer} simplified this design by routing each token to a single expert, improving training stability and efficiency while retaining scalability.

More recently, Time-MoE~\cite{shi2024timemoe} integrated sparse MoE layers into a decoder-only Transformer architecture for time-series foundation models, allowing scaling to billions of parameters while activating only a small subset of experts during inference. 
Liu \emph{et al.} proposed MOIRAI-MoE~\cite{liu2024mixture}, arguing that hand-crafted frequency-level specialization remains too coarse for non-stationary and heterogeneous time series. 
Instead, they rely on token-level routing within sparsely activated MoE layers while sharing input and output projections, allowing expert behaviors to emerge automatically. 
More recently, Time Tracker~\cite{liang2026timetracke} combined MoE Transformers with any-variate attention and learned inter-series graphs to capture temporal heterogeneity and multivariate dependencies.

The design of the gating mechanism plays a critical role in MoE performance. 
Traditional approaches based on dense softmax gating do not scale efficiently as the number of experts increases. 
Aviv \emph{et al.} proposed MoGU~\cite{aviv2025mogu}, a probabilistic MoE framework that incorporates uncertainty-aware gating by weighting expert contributions according to predictive variance. 
This strategy improves both point forecasting accuracy and uncertainty calibration in multivariate time-series settings.

Despite these advances, several challenges remain open, including expert interpretability, routing stability, online adaptation, and effective multivariate expert selection. 
Addressing these issues is essential for the reliable deployment of MoE-based forecasting models in real-world industrial and cyber--physical applications.

\section{Notation and Definitions}\label{sec3}

\begin{definition}[Time-Series Dataset]

Let $\mathcal{P} = \{\mathcal{P}_1, \mathcal{P}_2, \dots, \mathcal{P}_H\}$ denote a dataset composed of $H$ subsets of time series, where each subset

\begin{equation}
    \mathcal{P}_i = \{X_i^{(1)}, X_i^{(2)}, \dots, X_i^{(N_i)}\}
\end{equation}

contains $N_i$ time series acquired from repeated observations of the same physical process or system under comparable operating conditions.

Each time series $X_i^{(j)} \in \mathbb{R}^{L_i}$ is a real-valued univariate or multivariate signal of length $L_i$.  
Within a given subset $\mathcal{P}_i$, the time series are assumed to be independently and identically distributed (i.i.d.) realizations of an underlying stochastic process, while different subsets may correspond to distinct domains, sensing modalities, assets, or operating regimes.
\end{definition}

\begin{definition}[Neural Network]

A neural network is defined as a triple

\begin{equation}
\mathcal{N} = (\mathcal{S}, \mathcal{W}, \Phi),
\end{equation}

where:
\begin{itemize}
    \item $\mathcal{S} = \{S_1, S_2, \dots, S_K\}$ denotes the ordered set of network layers, with $S_1$ and $S_K$ representing the input and output layers, respectively, and the intermediate layers corresponding to hidden layers;
    \item $\mathcal{W} \subseteq \mathcal{S} \times \mathcal{S}$ denotes the set of directed inter-layer connections;
    \item $\Phi = \{\phi_k\}_{k=2}^{K}$ denotes the collection of activation functions associated with each non-input layer.
\end{itemize}

Each layer $S_k$ contains $s_k$ neurons $\{n_{k,1}, \dots, n_{k,s_k}\}$.  
For neuron $n_{k,l}$, let $u_{k,l}$ and $v_{k,l}$ denote its pre-activation input and output, respectively, such that

\begin{equation}
v_{k,l} = \phi_k(u_{k,l}), \quad 2 \leq k \leq K.
\end{equation}

The pre-activation input is given by

\begin{equation}
u_{k,l} = b_{k,l} + \sum_{j=1}^{s_{k-1}} w_{k-1,j,l} \, v_{k-1,j},
\end{equation}
where $w_{k-1,j,l}$ denotes the weight connecting neuron $n_{k-1,j}$ to neuron $n_{k,l}$, and $b_{k,l}$ is a bias term.

Let $\mathcal{D}_k$ denote the vector space associated with layer $S_k$.  
The neural network $\mathcal{N}$ thus defines a parametric mapping

\begin{equation}
h_{\mathcal{N}} : \mathcal{D}_1 \rightarrow \mathcal{D}_K,
\end{equation}
which is optimized for a given task through the minimization of a task-dependent loss function over the trainable parameters.
\end{definition}

\begin{definition}[Foundation Model for Time-Series Processing]
We define a foundation model for time-series processing (FM-TSP) as a modular neural architecture

\begin{equation}
\mathcal{F} = (\mathcal{E}, \mathcal{G}, \mathcal{T}),
\end{equation}
where:
\begin{itemize}
    \item $\mathcal{E} = \{E_1, E_2, \dots, E_M\}$ denotes a collection of $M$ pretrained representation encoders.  
    Each encoder $E_i$, implemented as a neural network, 
    maps an input time series, drawn from the datasets defined in Definition~1 and embedded in the input space $D_1$ to a latent representation space $\mathcal{D}_K ^{(i)}$ ($E_i: \mathcal{D}_1 \rightarrow \mathcal{D}_K ^{(i)}$).  
    The encoders are pretrained in a self-supervised manner, independently of any downstream task, with the objective of extracting transferable and task-agnostic representations from heterogeneous time-series data. 

    \item $\mathcal{G}$ denotes a generic aggregation and adaptation module responsible for integrating the latent representations produced by a selected subset of pretrained encoders.  
    This module operates in a shared feature space and captures cross-encoder relationships, enabling the fusion of heterogeneous representations into unified downstream features. Thus the mapping operated by the agregartion module is $\mathcal{G}: \{\mathcal{D}_K ^{(i)}\}_{i \in \mathcal{I}} \rightarrow \mathcal{D}_G$.

    \item $\mathcal{T} = \{\tau_1, \tau_2, \dots, \tau_T\}$ denotes a family of task-specific heads, where each $\tau_T$ corresponds to a model for a particular downstream task (e.g., forecasting, imputation, anomaly detection, or virtual sensing).  
    These task-specific models are lightweight and are trained or fine-tuned using limited labeled data such as $\tau_P: \mathcal{D}_G  \rightarrow \mathcal{D}_T$.
\end{itemize}

A gating mechanism is associated with the foundation model to dynamically select, for a given input time series, a subset of relevant encoders from $\mathcal{E}$.  
This conditional selection enables adaptive computation, allowing the FM-TSP to leverage encoder specialization while ensuring robustness, scalability, and generalization across diverse operating regimes.
\end{definition}

\section{Foundation Model for Time Series}\label{sec4}

\begin{figure}[t]
    \centering
    \includegraphics[width=3.4in]{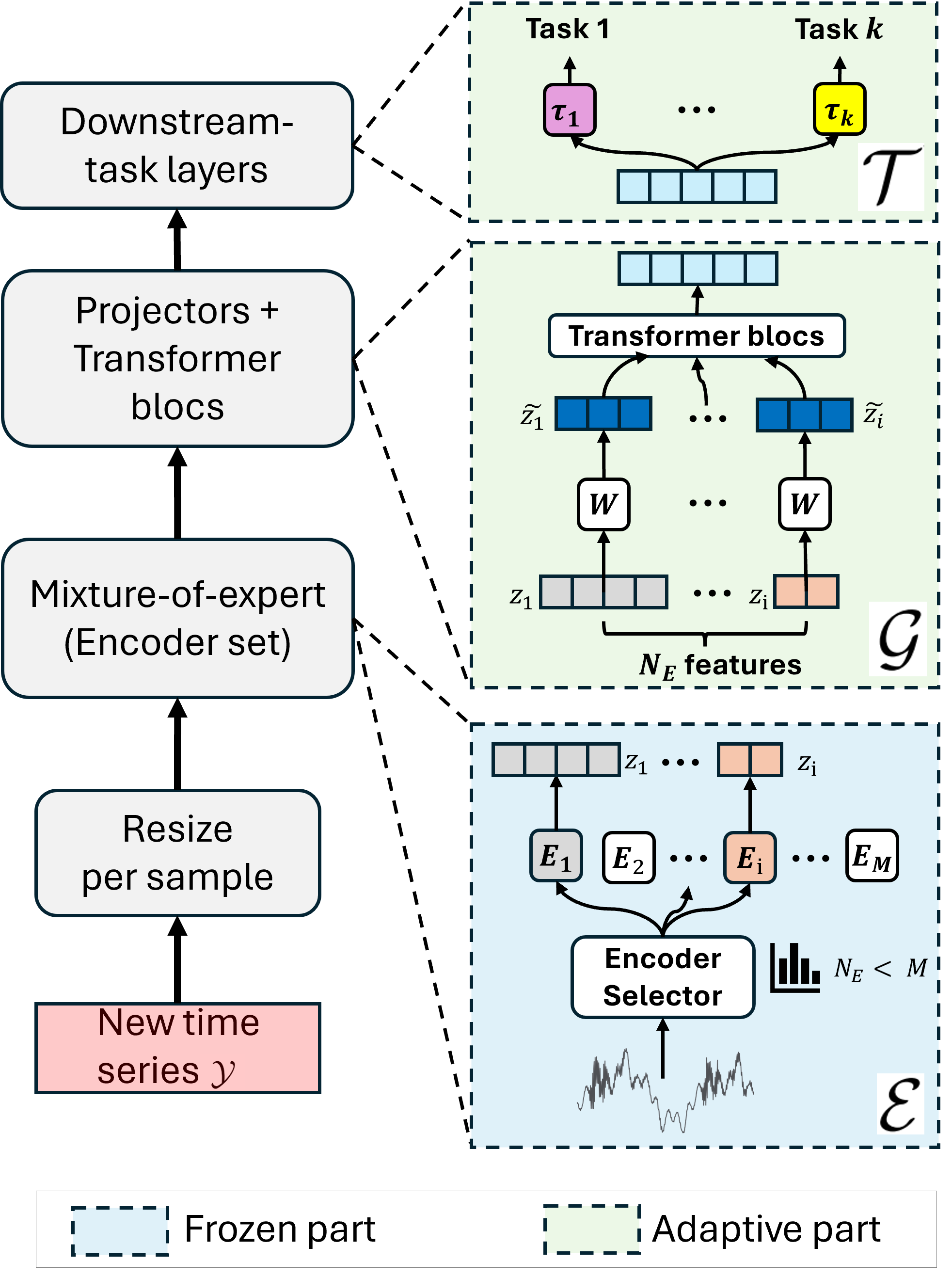}
    \caption{Conceptual framework of the proposed foundation model for time-series processing.}
    \label{fig:schema_Methodo}
\end{figure}

\subsection{General Framework}

Figure~\ref{fig:schema_Methodo} illustrates the overall architecture of the proposed foundation model for time-series processing (FM-TSP), denoted by

\begin{equation}
\mathcal{F} = (\mathcal{E}, \mathcal{G}, \mathcal{T}),
\end{equation}

in accordance with Definition~3. The proposed framework is designed to process time-series data drawn from the heterogeneous datasets introduced in Definition~1 by leveraging modular neural components formalized in Definition~2.

The FM-TSP is built upon a mixture of pretrained encoders $\mathcal{E} = \{E_i\}_{i=1}^{M}$, where each encoder $E_i$ is a neural network trained in a self-supervised manner on a specific subset $\mathcal{P}_i$ of the dataset collection $\mathcal{P}$. Each encoder implements a mapping

\begin{equation}
E_i : \mathcal{D}_1 \rightarrow \mathcal{D}_K^{(i)},
\end{equation}

where $\mathcal{D}_1$ denotes the common input space associated with raw or preprocessed time-series segments, and $\mathcal{D}_K^{(i)}$ denotes the latent representation space learned by encoder $E_i$. The encoders are pretrained independently of any downstream task and are subsequently frozen.

Given a new dataset $\mathcal{Y} = \{y_j\}_{j=1}^{N_{\mathcal{Y}}}$, not observed during pretraining, the FM-TSP operates according to the following processing pipeline.

\paragraph{Encoder gating and selection}
A gating mechanism dynamically selects a subset of $N_E$ relevant encoders from $\mathcal{E}$ based on (i) the similarity between $\mathcal{Y}$ and the pretraining datasets $\{\mathcal{P}_i\}$ and (ii) the structural consistency between input-space and latent-space similarity relationships. This selection step ensures that only encoders whose learned representations are well aligned with the target data distribution are activated, yielding a conditional and modular computation scheme consistent with Definition~3.

\paragraph{Projection into a shared latent space}
The latent representations produced by the selected encoders generally lie in heterogeneous spaces $\mathcal{D}_K^{(i)}$ with varying dimensionalities. To enable joint processing, each encoder output $z_i \in \mathbb{R}^{\ell_i}$ is mapped to a shared $d$-dimensional representation space $\mathcal{D}_P$ via a learnable linear projector

\begin{equation}
\tilde{z_i} = W_i z_i + b_i, \quad W_i \in \mathbb{R}^{d \times \ell_i}.
\end{equation}

The set of projected representations $\{\tilde{z_i}\}$ constitutes a collection of aligned feature tokens in a common semantic space.

\paragraph{Downstream aggregation and task inference}
The aggregation and adaptation module $\mathcal{G}$ processes the projected representations using a Transformer-based architecture. 
Each projected representation $\tilde{z_i} \in \mathcal{D}_P$ is treated as an input token, yielding a sequence

\begin{equation}
\tilde{Z} = [\tilde{z_1}, \dots, \tilde{z}_{N_E}] \in \mathbb{R}^{N_E \times d}.
\end{equation}

After an additional shared linear embedding and the addition of learnable positional encodings, the resulting token sequence is processed by a self-attention mechanism to capture cross-encoder relationships and dependencies. The output of $\mathcal{G}$ defines a task-agnostic representation in the space $\mathcal{D}_G$.

Finally, a task-specific head $\tau_T \in \mathcal{T}$ maps the aggregated representation to the output space associated with the downstream task of interest, such that

\begin{equation}
\tau_T : \mathcal{D}_G \rightarrow \mathcal{D}_T.
\end{equation}

During fine-tuning, only the parameters of the projectors, aggregation module, and task-specific head are updated, while the pretrained encoders remain fixed. This design enables efficient task adaptation using limited labeled data while preserving the generality of the pretrained representations.

\subsection{Encoder Pretraining}

The collection of encoders $\mathcal{E} = \{E_i\}_{i=1}^{M}$ is pretrained using self-supervised learning on the time-series datasets defined in Definition~1.  
Each encoder $E_i$ is associated with a specific training configuration, defined by a pair $(\mathcal{P}_j, \mathcal{N}_k)$, where $\mathcal{P}_j \subset \mathcal{P}$ denotes a subset of time-series data and $\mathcal{N}_k$ specifies the neural architecture and its hyperparameters, in the sense of Definition~2.  
This design encourages encoder specialization with respect to data characteristics, sensor modalities, or operating regimes.

For notational simplicity, the indices $(j,k)$ are omitted in the remainder of this section.  
Each encoder is pretrained independently and optimized to map input time-series segments from the common input space $\mathcal{D}_1$ to a latent representation space $\mathcal{D}_K^{(i)}$ without relying on task-specific supervision.

Let $X \in \mathcal{P}_j$ be a time series of length $L$, drawn from a selected dataset subset.  
The series is segmented into $N$ overlapping or non-overlapping windows $\{x_n\}_{n=1}^{N}$, with $x_n \in \mathbb{R}^{w}$ and $w \ll L$.  Noted that the choice of overlapping level depends on the nature of time-series.
Each segment is embedded in the input space $\mathcal{D}_1$ and mapped to a latent representation

\begin{equation}
z_n = E_i(x_n) \in \mathcal{D}_K^{(i)}.
\end{equation}

\begin{figure}[t]
    \centering
    \includegraphics[width=5in]{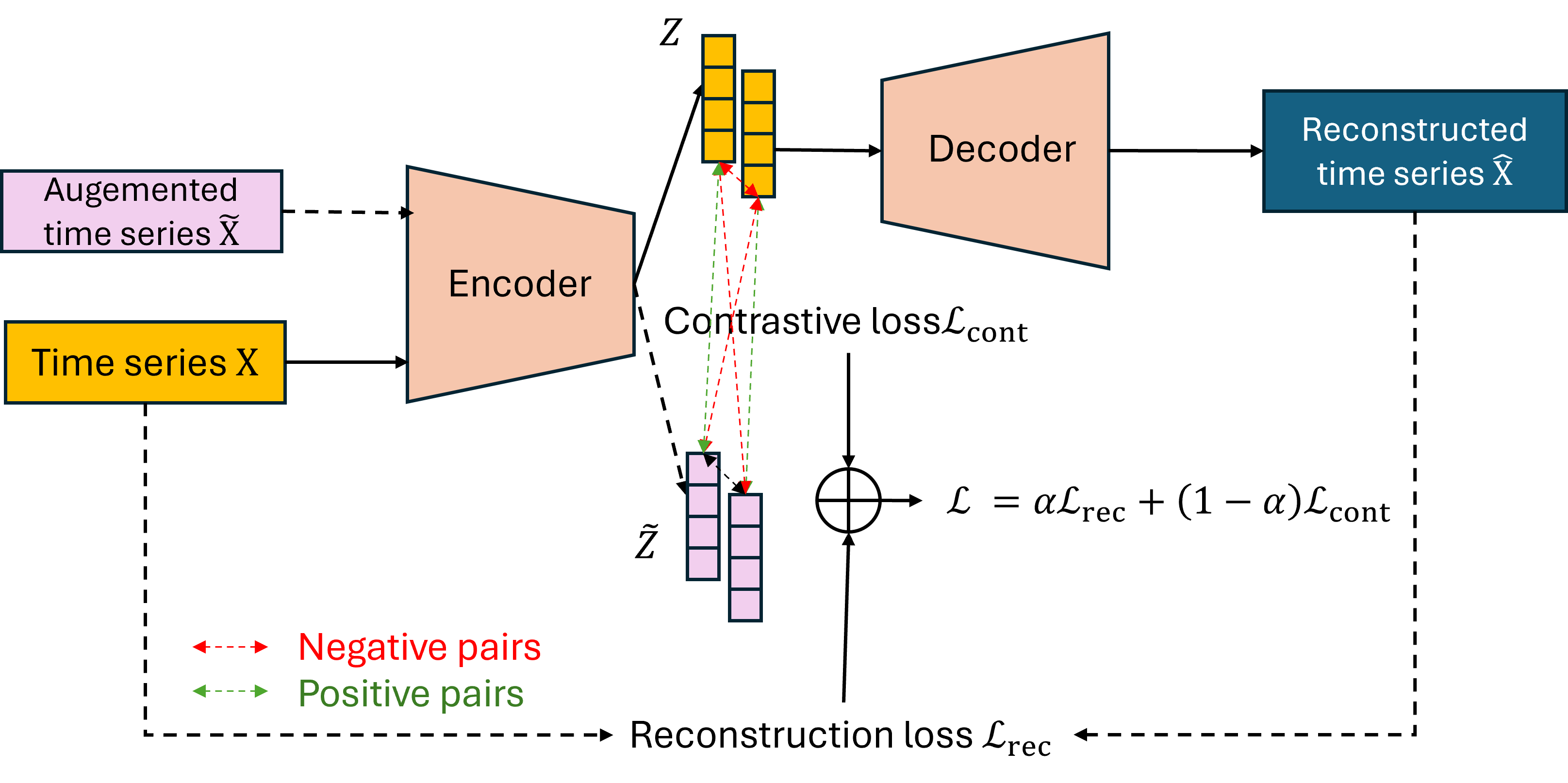}
    \caption{Encoder pretraining with a lightweight decoder to focus learning on the encoder.}
    \label{fig:pretrain}
\end{figure}

To concentrate representational capacity in the encoder, each pretraining pipeline uses an autoencoder structure with a lightweight decoder composed of two hidden layers. As illustrated by Figure~\ref{fig:pretrain}, two complementary self-supervised objectives are employed: reconstruction objective combined with the contrastive objective.

\subsubsection{Reconstruction objective}
The decoder reconstructs each input segment $\hat{x}_n$ from its latent representation $z_n$.  
The reconstruction loss is defined using the Huber loss \cite{ref:Huber}

\begin{equation}
\mathcal{L}_{\text{rec}} = 
\frac{1}{N}
\sum_{n=1}^{N}
\begin{cases}
\frac{1}{2}(x_n-\hat{x}_n)^2,
& \text{if } |x_n-\hat{x}_n| \le \delta, \\[6pt]
\delta \left( |x_n-\hat{x}_n| - \frac{\delta}{2} \right),
& \text{if } |x_n-\hat{x}_n| > \delta.
\end{cases}
\end{equation}

which provides robustness to outliers and measurement noise commonly encountered in industrial time-series data.

\subsubsection{Contrastive objective}
To regularize the latent space and enforce invariance to partial corruption, a contrastive learning objective is added.  
For each segment $x_n$, a set of $m$ masked augmentations  
\begin{equation}
    \mathcal{X}_n = \{\mathcal{M}^{(q)} (x_{n})\}_{q=0}^{m} 
\end{equation}

is generated by randomly masking contiguous portions of the signal.
Note that for $q = 0$, the masking operator reduces to the identity mapping, i.e.,
$x_n = \mathcal{M}^{(0)}(x_n)$.
The corresponding latent representations are denoted by: 
\begin{equation}
    z_n^{(q)}=E_i \big(\mathcal{M}^{(q)} (x_{n}) \big)  \quad, \quad  \mathcal{Z}_n = \{ z_n^{(q)} \}_{q=0}^{m}
\end{equation}

Thus, we define positive and negative pairs as:

\begin{itemize}
    
    \item Positive: $(z_n^{(q)},\, z_n^{(q')})$ with $z_n^{(q)}, z_n^{(q')} \in \mathcal{Z}_n$ and $q \neq q'$.
    \item Negative: $(z_n^{(q)}, z_k^{(q')})$ with
$z_n^{(q)} \in \mathcal{Z}_n$, $z_k^{(q')} \in \mathcal{Z}_k$, and $k \neq n$.
\end{itemize}

Let $\mathcal{Z} = \bigcup_{n=1}^N \mathcal{Z}_n$ denote the set of all masked latent representations in a mini-batch.
For each anchor representation $z \in \mathcal{Z}_n$, we define the corresponding positive set as
$\mathcal{Z}^+ = \mathcal{Z}_n \setminus \{z\}$.
The contrastive loss uses the NT-Xent formulation \cite{ref:NTXent}:

\begin{equation}
    \mathcal{L}_{\text{cont}}
    = - \sum_{z \in \mathcal{Z}} \sum_{z' \in \mathcal{Z}^+}
    \log \frac{\exp(\mathbf{R}_{z,z'} / \tau)}
    {\sum_{z'' \in \mathcal{Z} \setminus \{z\}}
    \exp(\mathbf{R}_{z,z''} / \tau)},
\end{equation}

where $\tau$ is the temperature and $\mathbf{R} = \text{Sim}(\mathcal{Z})$ is the cosine-similarity matrix with entries
$\mathbf{R}_{u,v} = \frac{u^\top v}{\lVert u \rVert \, \lVert v \rVert}$.

\subsubsection{Pretraining loss}
The final self-supervised training objective for encoder $E_i$ is a weighted combination of the reconstruction and contrastive losses:

\begin{equation}
\mathcal{L}_{\text{pre}} = \alpha \, \mathcal{L}_{\text{rec}} + (1 - \alpha)\, \mathcal{L}_{\text{cont}},
\label{equ:CombContrRec}
\end{equation}

where $\alpha \in [0,1]$ controls the trade-off between signal fidelity and representation invariance.

After pretraining, all encoder parameters are frozen.  
The resulting encoders provide stable and task-agnostic representations that are subsequently reused by the aggregation and adaptation module $\mathcal{G}$ for downstream tasks, as described in Section~IV-A.

%------------------------
\subsection{Encoder Gating and Selection}

After self-supervised pretraining, the objective of the gating mechanism is to identify, for a given target dataset, the subset of pretrained encoders whose learned representations are most relevant.  
This selection step plays a central role in the proposed FM-TSP, as it enables conditional computation at the encoder level, in accordance with Definition~3.

Let $\mathcal{E} = \{E_i\}_{i=1}^{M}$ denote the collection of pretrained encoders, where each encoder $E_i$ has been trained on a dataset subset $\mathcal{P}_i$ as defined in Section~IV-B.  
Let $\mathcal{Y} = \{y_n\}_{n=1}^{N_{\mathcal{Y}}}$ denote a target dataset drawn from the same input space $\mathcal{D}_1$ but not observed during pretraining.  
The goal of the gating mechanism is to select a subset $\mathcal{E}^\star \subset \mathcal{E}$ of $N_E$ encoders that are best aligned with the statistical and structural properties of $\mathcal{Y}$.

Encoder selection is performed using a two-stage procedure based on complementary criteria: data similarity and structural correlation.

\subsubsection{Data Similarity Criterion}

The first criterion evaluates the similarity between the target dataset $\mathcal{Y}$ and the pretraining dataset $\mathcal{P}_i$ associated with encoder $E_i$.  
The underlying assumption (\textbf{H1}) is that encoders pretrained on data distributions closer to the target domain are more likely to extract informative and stable representations.

Rather than comparing raw time series directly, each time-series segment is mapped to a fixed-dimensional feature representation that captures its statistical and spectral characteristics. 
This approach relies on assumption (\textbf{H2}), which states that statistical and spectral features are sufficient to represent the stochastic process associated with each time series.
Let $\phi(\cdot)$ denote an embedding operator
\begin{equation}
\phi : \mathbb{R}^{w} \rightarrow \mathbb{R}^{N_\phi},
\end{equation}

where $w$ is the segment length and $N_\phi$ the embedding dimension.

The embedding $\phi(\cdot)$ concatenates:
\begin{itemize}
    \item \emph{Statistical descriptors}, summarizing amplitude distribution, temporal variability, and local dynamics (e.g., moments, quantiles, autocorrelations, event rates). Such heterogeneous descriptors have proven effective for capturing broad behavioural classes of time series \cite{fulcher_highly_2013,lubba_catch22_2019};
    \item \emph{Spectral descriptors}, derived from the magnitudes of low-frequency Fourier coefficients, complemented by spectral centroid and roll-off indices. These descriptors provide a compact representation of dominant periodicities, smoothness, and frequency distribution, widely used in signal-processing contexts \cite{tzanetakis_musical_2002}.
\end{itemize}

Let
\begin{equation}
\Phi_{\mathcal{Y}} = \{\phi(y_n)\}_{n=1}^{N_{\mathcal{Y}}} \in \mathbb{R}^{N_{\mathcal{Y}} \times N_\phi}
\end{equation}

and 
\begin{equation}
\Phi_{\mathcal{P}_i} = \{\phi(x_m)\}_{x_m \in \mathcal{P}_i} \in \mathbb{R}^{N_i \times N_\phi}
\end{equation}

denote the embedded representations of the target dataset and the pretraining dataset associated with encoder $E_i$, respectively.

A dataset-level dissimilarity score is computed as the mean pairwise Euclidean distance
\begin{equation}
d(\mathcal{P}_i, \mathcal{Y}) = \frac{1}{N_i N_{\mathcal{Y}}} \sum_{m=1}^{N_i} \sum_{n=1}^{N_{\mathcal{Y}}}
\left\lVert \phi(y_n) - \phi(x_m) \right\rVert_2.
\end{equation}

Encoders are ranked in ascending order of $d(\mathcal{P}_i, \mathcal{Y})$, and only the top-ranked candidates are retained for the second selection stage.  
More computationally intensive alternatives, such as distances based on optimal transport, can be employed when feasible \cite{opt_transport}.

\subsubsection{Structural Correlation Criterion}

The second criterion assesses whether an encoder preserves the local similarity structure of the target data when mapping it from input space $\mathcal{D}_1$ to its latent representation space $\mathcal{D}_K^{(i)}$.  
This step is motivated by assumption (\textbf{H3}), complementary to (\textbf{H2}), which states that a suitable encoder should maintain relative neighborhood relationships across representation spaces.

For a given encoder $E_i$, each element $y_n \in \mathcal{Y}$ is mapped to a latent vector
\begin{equation}
z_n = E_i(y_n).
\end{equation}
Let
\begin{equation}
R_{\mathcal{Y}} \in \mathbb{R}^{N_{\mathcal{Y}} \times N_{\mathcal{Y}}},
\quad
R_Z^{(i)} \in \mathbb{R}^{N_{\mathcal{Y}} \times N_{\mathcal{Y}}}
\end{equation}

denote the pairwise cosine similarity matrices computed in the input space and the latent space associated with encoder $E_i$, respectively. The structural alignment score is defined as the normalized Pearson correlation between $R_{\mathcal{Y}}$ and $R_Z^{(i)}$:

\begin{equation}
r(E_i, \mathcal{Y}) = \frac{1}{2} \left( 1 + 
\operatorname{corr}\!\left(\operatorname{vec}(R_{\mathcal{Y}}), \operatorname{vec}(R_Z^{(i)}) \right) \right)
\end{equation}

where $\operatorname{vec}(\cdot)$ stacks matrix entries into a vector.  
The normalization maps the score to the interval $[0,1]$.

Finally, the $N_E$ encoders with the highest structural correlation scores are selected to form the active encoder subset $\mathcal{E}^\star$, which is subsequently used by the projection and aggregation modules described in Section~IV-A.
%------------------------

%-------------------------
\subsection{Fine-Tuning for the Downstream Task}

After the encoder gating and selection stage, the selected subset of encoders $\mathcal{E}^\star \subset \mathcal{E}$ is used to adapt the foundation model to a specific downstream task.  
In this phase, all pretrained encoders $E_i \in \mathcal{E}^\star$ are kept frozen, while the adaptive components of the model are trained, namely the projection modules, the aggregation module, and the task-specific head. This design preserves the generality of the pretrained representations while enabling efficient task adaptation using limited labeled data.

Let $\tau_T \in \mathcal{T}$ denote a downstream task, with associated target dataset
$\mathcal{Y} = \{(y_n, \ell_n)\}_{n=1}^{N_p}$,
where $y_n \in \mathcal{D}_1$ and $\ell_n$ denotes the task-specific target.

For each input $y_n$, the selected encoders produce latent representations
\begin{equation}
z_n^{(i)} = E_i(y_n), \quad E_i \in \mathcal{E}^\star.
\end{equation}

\subsubsection{Projection and alignment}
Since the encoder outputs $z_n^{(i)} \in \mathcal{D}_K^{(i)}$ generally lie in heterogeneous latent spaces with different dimensionalities, each representation is first mapped to a shared $d$-dimensional space $\mathcal{D}_P$ using a learnable linear projector:
\begin{equation}
\tilde{z}_n^{(i)} = P_i(z_n^{(i)}) = W_i z_n^{(i)} + b_i,
\quad W_i \in \mathbb{R}^{d \times \ell_i}.
\end{equation}

To encourage semantic alignment across projected representations originating from different encoders, an early-stage regularization term based on Maximum Mean Discrepancy (MMD) is introduced, which is  used for measuring the difference between two probability distributions \cite{gretton_kernel_2012}.  
Let $\mathbb{P}_i$ denote the empirical distribution of the projected features $\{\tilde{z}_n^{(i)}\}_{n=1}^{N_p}$ produced by projector $P_i$.  
The alignment regularization loss is defined as
\begin{equation}
\mathcal{L}_{\text{reg}} =
\frac{2}{N_E (N_E - 1)}
\sum_{1 \leq i < j \leq N_E}
\operatorname{MMD}^2(\mathbb{P}_i, \mathbb{P}_j),
\end{equation}
where $\operatorname{MMD}(\cdot,\cdot)$ denotes the kernel-based two-sample distance.

\subsubsection{Aggregation and task inference}
The projected representations $\{\tilde{z}_n^{(i)}\}$ are then aggregated by the generic module $\mathcal{G}$.  
Each projected feature is treated as a token in a sequence processed by a Transformer-based encoder, enabling self-attention across encoder representations and capturing cross-expert dependencies.  
The output of $\mathcal{G}$ defines a task-agnostic representation $h_n \in \mathcal{D}_G$.

A task-specific head $\tau_T$ maps the aggregated representation to the task output space:
\begin{equation}
\hat{\ell}_n = \tau_T(h_n), \quad \tau_T : \mathcal{D}_G \rightarrow \mathcal{D}_T.
\end{equation}

\subsubsection{Optimization objective (optional)}
The fine-tuning loss is defined as a combination of the task-specific loss $\mathcal{L}_{\tau_T}$ and the alignment regularization term:

\begin{equation}
    \mathcal{L}_{\text{ft}} =
\begin{cases}
\mathcal{L}_{\tau_T} + \beta \mathcal{L}_{\text{reg}}, & \text{if epoch} \leq n_e, \\
\mathcal{L}_{\tau_T}, & \text{otherwise}.
\end{cases}
\label{eq_reg}
\end{equation}

where $\beta$ controls the strength of the alignment constraint and $n_e$ denotes the number of warm-up epochs during which the regularization is applied.  
This schedule ensures that early training stages promote representation compatibility, while later stages prioritize task-specific performance.

During fine-tuning, only the parameters of the projectors $\{P_i\}$, the aggregation module $\mathcal{G}$, and the task-specific head $\tau_T$ are updated, whereas all encoders in $\mathcal{E}^\star$ remain fixed.

\section{Ablation Study}\label{sec5}

This section evaluates the individual contributions of the main components of the proposed foundation model for time-series processing (FM-TSP).  
All experiments are designed to isolate the effect of each architectural and methodological choice while keeping the remaining components unchanged.  
Unless stated otherwise, all configurations share identical training protocols, data splits, and optimization settings to ensure fair comparison. For all tests, multiple tasks, including forecasting, imputation, and reconstruction, are evaluated across different data types. To facilitate comparison, we present box plots illustrating the distribution of performance metrics. Three evaluation metrics are considered: mean squared error (MSE), dynamic time warping (DTW), and the Jacobian norm (JAC). While MSE and DTW assess the similarity between the model outputs and the original signals, the Jacobian-based metric is introduced to quantify the sensitivity of the model \cite{TS_Park2024, TS_Hoffman2019, TS_Liu2024, TS_Jakubovitz2018}. Specifically, a low Jacobian value suggests that the model may not have sufficiently learned the underlying data patterns, whereas an excessively high Jacobian value indicates that the model is overly sensitive to small input variations, potentially reflecting overfitting.
\subsection{Experimental Setup}

To promote encoder diversity during the pretraining phase, multiple factors are varied when constructing the encoder collection $\mathcal{E}$.  
Each encoder configuration is characterized by the following factors:

\begin{itemize}
    \item \textbf{Pretraining datasets} $\mathcal{P}_i$, drawn from heterogeneous domains and sensing modalities as defined in Definition~1;
    \item \textbf{Encoder architecture}, including MLP, CNN, and Transformer-based networks as formalized in Definition~2;
    \item \textbf{Signal masking ratio}, controlling the degree of temporal corruption applied during self-supervised pretraining;
    \item \textbf{Contrastive loss weight} $\alpha$ (see Eq.~\ref{equ:CombContrRec}), defining the trade-off between reconstruction and contrastive objectives;
    \item \textbf{Other hyperparameters} of training phase such as learning rate (see Table~\ref{tab:encoder_factors});
\end{itemize}

\begin{table*}[t]
\caption{Factors characterizing the pretrained encoder collection}
\label{tab:encoder_factors}
\centering
\renewcommand{\arraystretch}{1.15}
\begin{tabular}{|c|c|p{3.8cm}|p{4.2cm}|}
\hline
\textbf{Category} & \textbf{Factor} & \textbf{Description} & \textbf{Values} \\
\hline
\multirow{2}{*}{Data diversity}
& Pretraining dataset 
& Time-series subsets drawn from heterogeneous domains and sensing modalities (see Definition~1) 
& Vibration, acoustic, electricity, ECG, temperature, EEG, hydraulic pressure \\
\cline{2-4}
& Masking ratio 
& Proportion of signal randomly masked during self-supervised pretraining 
& 0.0, 0.25, 0.5 \\
\hline
\multirow{2}{*}{Model diversity}
& Encoder architecture 
& Neural network architectures implementing the encoder mapping (Definition~2) 
& MLP, CNN, Transformer \\
\cline{2-4}
& Latent dimension 
& Dimensionality of the encoder output space $\mathcal{D}_K^{(i)}$ 
& 8, 16, 32, 64 \\
\hline
\multirow{2}{*}{Learning objectives}
& Contrastive weight $\alpha$ 
& Trade-off coefficient between reconstruction and contrastive losses (Eq.~(15)) 
& 0.0, 0.25, 0.5, 0.75, 1.0 \\
\cline{2-4}
& Reconstruction loss 
& Robust reconstruction objective applied during pretraining 
& Huber loss \\
\hline
\multirow{2}{*}{Optimization}
& Initial learning rate 
& Learning rate at the beginning of pretraining 
& 0.05, 0.01, 0.005 \\
\cline{2-4}
& Learning rate decay 
& Multiplicative decay factor applied upon validation stagnation 
& 1.0, 0.5, 0.1 \\
\hline
\end{tabular}
\end{table*}
The values explored for these parameters are summarized in Table~\ref{tab:encoder_factors}.  
While additional factors could further enrich the encoder bank, this set provides sufficient diversity to validate the proposed framework while maintaining a controlled experimental scope.

\subsection{Effect of Self-Supervised Pretraining}

The first ablation evaluates the impact of self-supervised pretraining on downstream performance.  
We compare three encoder initialization strategies:

\begin{enumerate}
    \item encoders pretrained using reconstruction loss only;
    \item encoders pretrained using the combined reconstruction–contrastive objective proposed in Section~IV-B.
\end{enumerate}

\begin{figure}[h]
    \centering
    \includegraphics[width=5in]{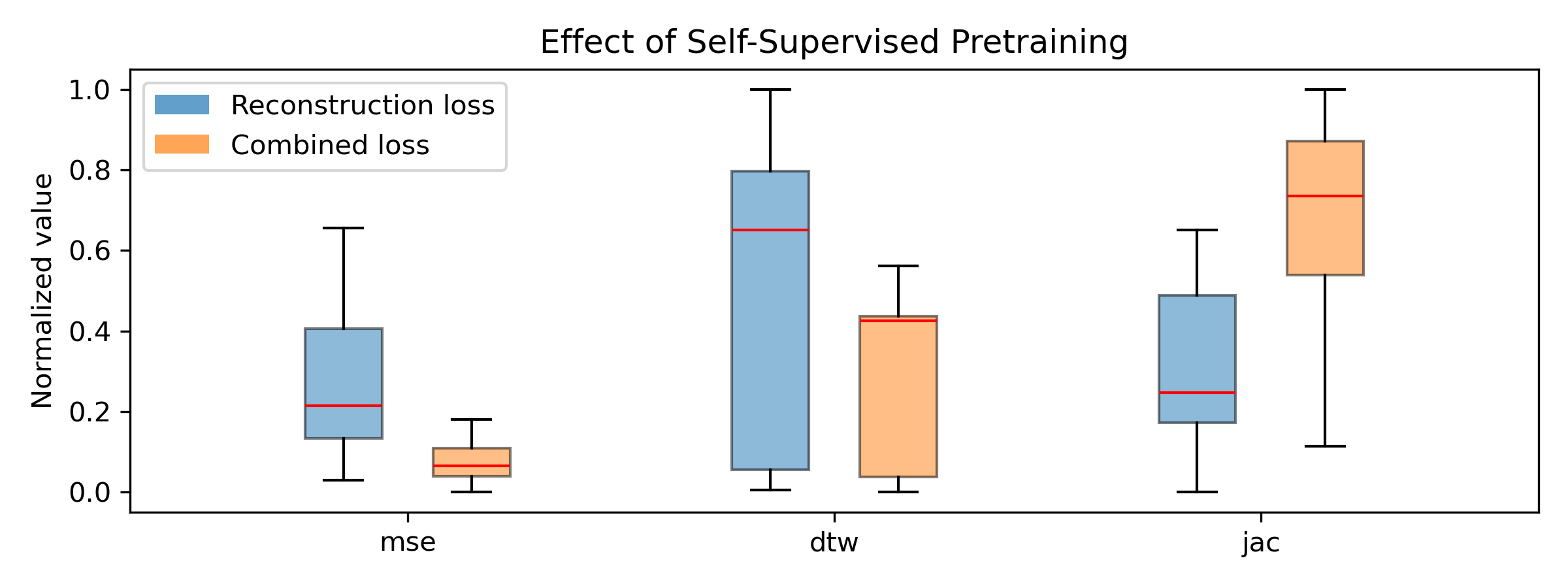}
    \caption{Effect of Self-Supervised Pretraining}
    \label{fig:Effect_SS}
\end{figure}

All subsequent pipeline components are kept identical. The results in Figure~\ref{fig:Effect_SS} show that combining reconstruction and contrastive objectives produces the most stable and accurate performance, highlighting the critical role of self-supervised representation quality in successful downstream adaptation.

\subsection{Impact of Encoder Gating and Selection}

The second ablation examines the role of the encoder gating and selection mechanism described in Section~IV-C.  
We evaluate three configurations:

\begin{enumerate}
    \item \textbf{Random selection}: A subset of encoders is randomly sampled, and this selection process is repeated 10 times to ensure diversity;
    \item \textbf{Worst-k selection}: the low ranked encoders are selected using the two-stage procedure based on data similarity and structural correlation;
    \item \textbf{Top-k selection}: the high ranked encoders are selected using the two-stage procedure based on data similarity and structural correlation;
\end{enumerate}

\begin{figure}[h]
    \centering
    \includegraphics[width=5in]{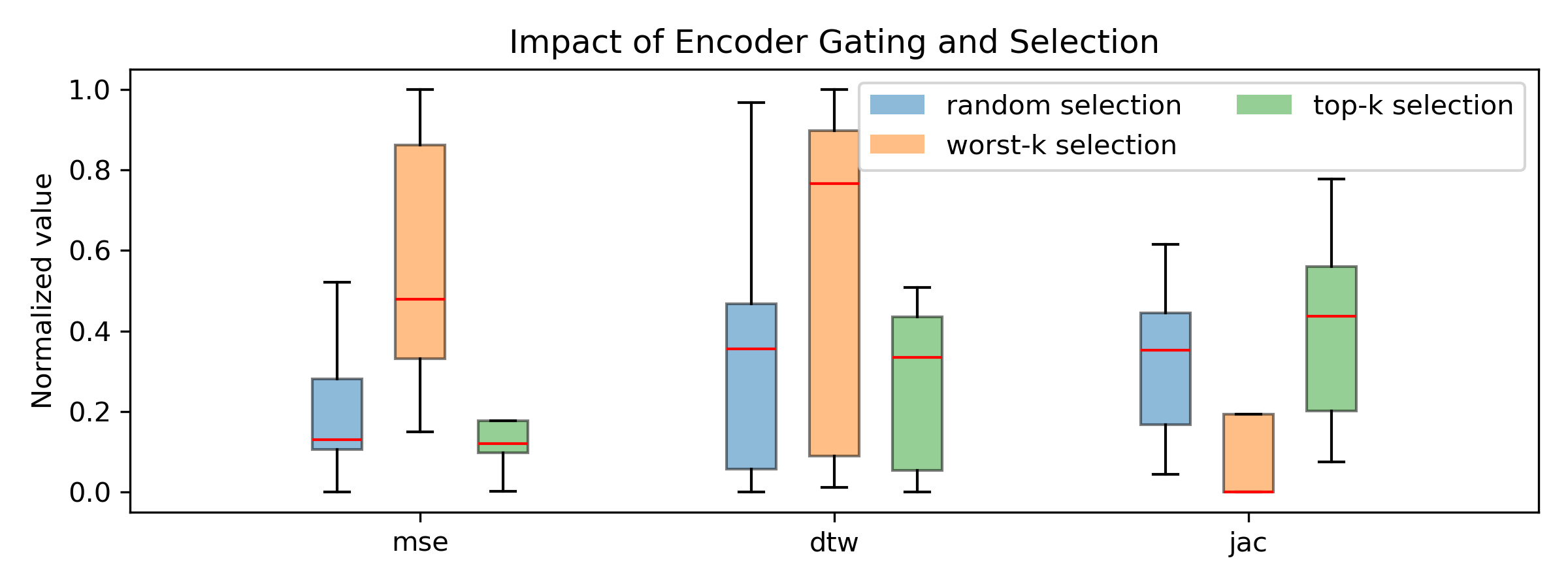}
    \caption{Impact of Encoder Gating and Selection}
    \label{fig:ImpactGating}
\end{figure}

Figure~\ref{fig:ImpactGating} shows that random selection yields high variance and inconsistent performance, whereas worst-k selection leads to the poorest results. The near-zero Jacobian-based values in the latter case indicate that the selected encoders do not learn meaningful representations. In contrast, the strong performance of top-k selection confirms that the proposed gating strategy consistently improves both accuracy and robustness, emphasizing the necessity of informed encoder selection rather than naive aggregation.

Although simple, this selection strategy is designed to remain computationally tractable and does not require joint retraining. More advanced selection criteria, such as optimal transport or learned meta-selection, are left for future work.

\subsection{Role of Projection and Representation Alignment}

This ablation studies the contribution of the projection and alignment mechanisms introduced in Section~IV-D.  
We compare the proposed approach to a baseline where encoder outputs are directly concatenated without projection or alignment regularization. Several $\beta$ (see Eq. \ref{eq_reg}) are also tested. 

\begin{figure}[h]
    \centering
    \includegraphics[width=5in]{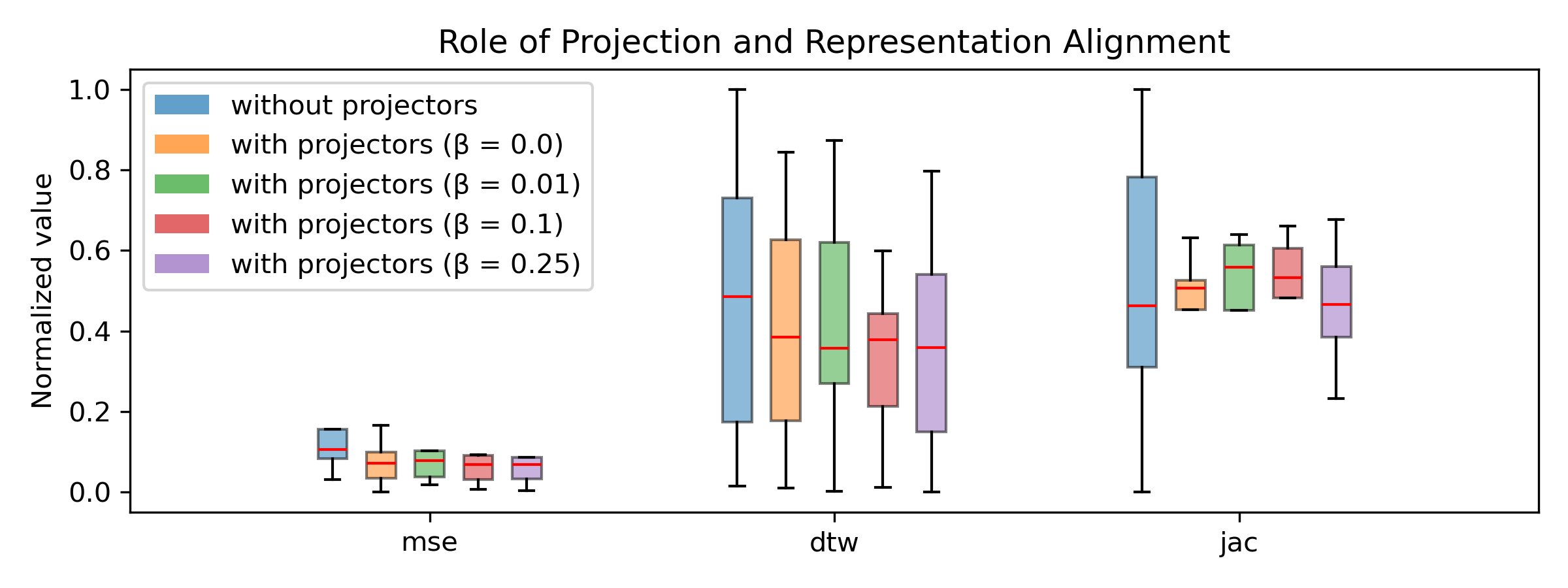}
    \caption{Role of Projection and Representation Alignment}
    \label{fig:RoleProjectionAlignement}
\end{figure}

As shown in Figure~\ref{fig:RoleProjectionAlignement}, direct concatenation without a projection module leads to unstable training dynamics and inferior performance. Introducing an alignment regularization term controlled by the coefficient $\beta$ (see Eq. \ref{eq_reg}) appears to improve stability, as evidenced by the slightly reduced variance in the box plots compared to the case where $\beta = 0$. This suggests that early-stage Maximum Mean Discrepancy (MMD) regularization facilitates convergence by aligning heterogeneous representations into a shared semantic space, thereby stabilizing performance across multiple trials in this ablation study. However, the average performance does not show significant improvement. This behavior may depend on the choice of $\beta$, for which no principled selection strategy is currently available beyond empirical tuning.

\subsection{Impact of Downstream aggregation}
The final ablation study evaluates the role of the downstream aggregation module $\mathcal{G}$ in combining the heterogeneous representations produced by the selected encoders.  
While the projection and alignment stages ensure that encoder outputs lie in a shared feature space, the aggregation mechanism determines how cross-encoder interactions are modeled before task-specific inference.

We compare the proposed Transformer-based aggregation module with simpler aggregation strategies commonly used in representation learning:
\begin{itemize}
    \item \textbf{Concatenation}: projected encoder outputs are concatenated and directly fed to the task-specific head;
    \item \textbf{Mean pooling}: projected representations are averaged across encoders, yielding a single aggregated feature vector;
    \item \textbf{Proposed aggregation}: a self-attention-based Transformer encoder that treats each projected encoder output as an individual token.
\end{itemize}

\begin{figure}[h]
    \centering
    \includegraphics[width=5in]{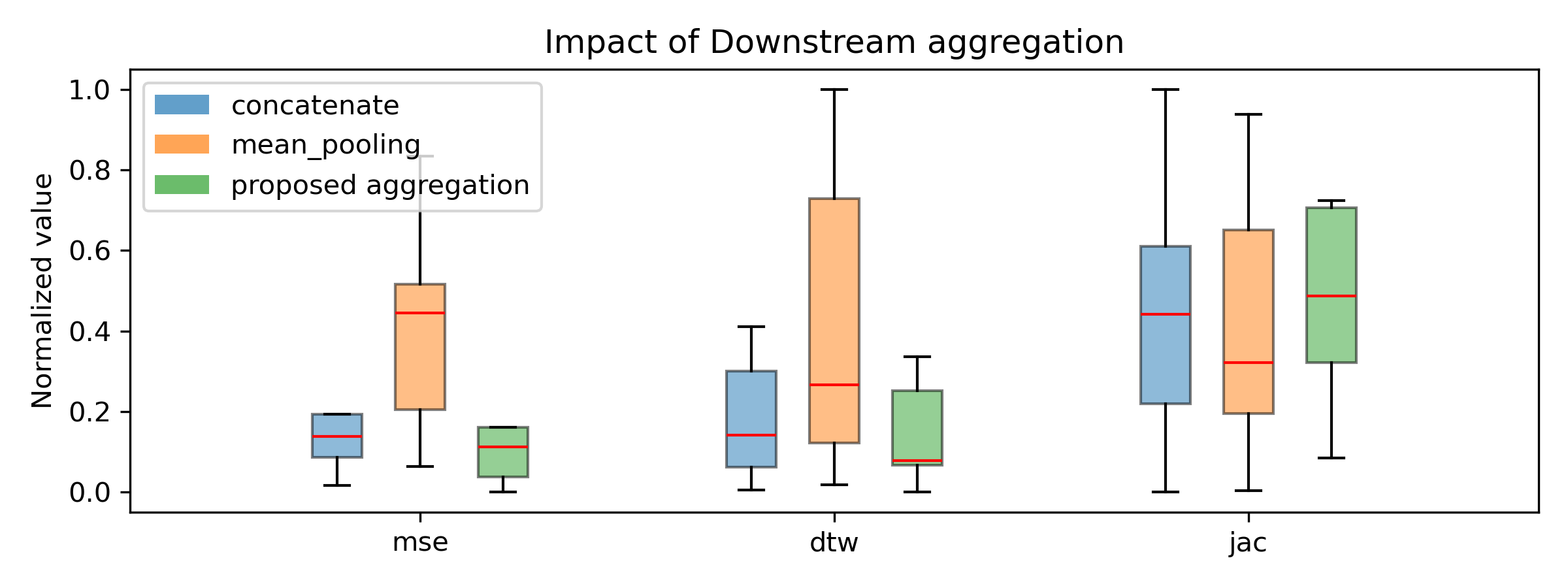}
    \caption{Impact of Downstream aggregation}
    \label{fig:ImpactAggregation}
\end{figure}
All configurations share the same encoder selection, projection, and fine-tuning settings, and differ only in the aggregation strategy.  
Experimental results, summarized in Fig \ref{fig:ImpactAggregation}, show that naive aggregation schemes such as concatenation or mean pooling lead to suboptimal performance and high variance (especially in case of mean pooling). In contrast, the Transformer-based aggregation consistently achieves superior results across tasks.

These observations highlight the importance of explicitly modeling cross-encoder relationships rather than assuming uniform or independent contributions.  
Self-attention enables the aggregation module to dynamically reweight encoder representations based on their relevance to the target task, capturing complementary patterns while suppressing redundant or noisy features. This adaptive behavior is especially beneficial in non-stationary and multi-regime time-series settings, which are typical of engineering digital twin applications.

Overall, this ablation confirms that downstream aggregation is a critical component of the proposed FM-TSP.  
The performance gains obtained with the Transformer-based aggregation cannot be reproduced by simple pooling operations, underscoring the necessity of a learnable and context-aware aggregation mechanism to fully exploit the diversity of pretrained encoders.

\subsection{Discussion}
%-----------------------
The ablation studies presented in this section provide a comprehensive evaluation of the key design choices underlying the proposed foundation model for time-series processing (FM-TSP).  
Taken together, these experiments highlight the necessity of jointly addressing representation learning, model selection, alignment, and aggregation to achieve robust and generalizable performance across diverse downstream tasks. Overall, these findings indicate that FM-TSP derives its strength not from any isolated component, but from the coordinated interaction of self-supervised representation learning, principled encoder selection, explicit alignment, and context-aware aggregation.

\section{Case study}\label{sec6}
The ablation studies presented in Section~V systematically validate the individual design choices underlying the proposed FM-TSP architecture, demonstrating the necessity of self-supervised pretraining, principled encoder gating, and explicit representation alignment.  
While these controlled experiments provide strong evidence of the internal consistency and effectiveness of the proposed framework, they do not fully capture its behavior under realistic deployment conditions.

We therefore complement the ablation analysis with a set of representative case studies designed to evaluate the proposed model in both standardized benchmarking settings and real-world industrial scenarios.  
These studies assess the ability of FM-TSP to generalize across tasks, data regimes, and operational contexts, thereby validating its practical relevance for engineering digital twin applications.

Two main case studies are proposed: 
\begin{itemize}
    \item The open-source ETT dataset \cite{zhou2021informer} is used to evaluate several common tasks, including imputation, long-term forecasting, and few-shot forecasting. This case study serves as a benchmark for localizing the performance of our model.
    \item A real-world industrial task related to virtual sensor is considered to further assess the model in practical scenario.
\end{itemize}

\subsection{Benchmark case study}

\subsubsection{Dataset description}

The benchmark evaluation is conducted using the publicly available Electricity Transformer Temperature (ETT) dataset, which has been widely adopted for assessing time-series forecasting and imputation models \cite{zhou2021informer}.  
The dataset contains multivariate time-series recordings collected from electricity transformers, capturing both load-related variables and environmental conditions over extended periods.

We consider four standard subsets of the ETT dataset, namely ETTh1, ETTh2, ETTm1, and ETTm2.  
The subsets differ in both temporal resolution and operating periods: ETTh1 and ETTh2 are sampled at an hourly frequency, while ETTm1 and ETTm2 are sampled at a finer time resolution.  
Each subset consists of multiple correlated channels, including power load measurements, oil temperature, and auxiliary variables, reflecting distinct operating regimes and seasonal variations.

From a formal perspective, each ETT subset corresponds to a target dataset $\mathcal{Y} \not\subset \mathcal{P}$ as defined in Definition~1.  
The resulting time-series are multivariate, non-stationary, and exhibit regime-dependent dynamics, making them well suited for evaluating the robustness and generalization capabilities of the proposed FM-TSP framework.

The ETT datasets naturally support multiple downstream tasks using a shared data source.  
In this work, they are leveraged to evaluate imputation, long-term forecasting, and few-shot forecasting under controlled and reproducible conditions, enabling direct comparison with existing time-series models while remaining aligned with real-world digital twin perception scenarios.

\subsubsection{Downstream Tasks}

To assess the versatility and generalization capabilities of the proposed foundation model for time-series processing (FM-TSP), we evaluate its performance on three representative downstream tasks using the ETT benchmark datasets.  
These tasks are selected to cover complementary aspects of time-series perception commonly encountered in digital twin applications, including reconstruction, prediction, and adaptation under data scarcity.

\paragraph{Imputation}
The imputation task aims to recover missing or corrupted values in multivariate time-series.  
Given an input sequence in which a fraction of observations is masked, the objective is to accurately reconstruct the missing entries by exploiting temporal dependencies and cross-variable correlations. Masking ratios of $\{0.125, 0.25, 0.375\}$ are considered in this case study. This task evaluates the ability of FM-TSP to leverage pretrained representations for signal reconstruction and noise robustness, which is particularly relevant in industrial monitoring scenarios where sensor failures or communication losses are frequent.

\paragraph{Long-term Forecasting}
The long-term forecasting task consists in predicting future time-series values over extended horizons based on a fixed-length historical context.  
Following standard ETT evaluation protocols, an input sequence of length 96 is used to forecast the next $\{96, 192, 336\}$ time steps.  
This task assesses the capability of FM-TSP to model long-range temporal dependencies and regime-dependent dynamics, while maintaining stability and accuracy as the prediction horizon increases.

\paragraph{Few-shot Forecasting}
In the few-shot forecasting setting, the forecasting task is performed under constrained supervision, where only a small fraction of the available labeled data is used for fine-tuning.  
Specifically, training subsets corresponding to $\{5\%, 10\%\}$ of the original training set are considered.  
This task directly evaluates the effectiveness of the pretrained encoders and the lightweight adaptation strategy, and reflects a common digital twin deployment scenario in which labeled data are scarce or costly to obtain.

Together, these three tasks provide a comprehensive evaluation of FM-TSP across reconstruction, prediction, and data-efficient adaptation regimes.  
By relying on a shared pretrained encoder set and task-specific lightweight heads, this experimental setup highlights the foundation-model paradigm adopted in this work, where a single representation backbone supports multiple heterogeneous downstream objectives.

\subsubsection{Data splits and Evaluation metrics}
Each ETT subset is divided into training, validation, and test segments a ratio of 70:10:20, respectively. Models are trained on the training set, hyperparameters are selected using the validation set, and final performance is reported on the held-out test set. This protocol is applied consistently for all downstream tasks and all considered baselines.

Model performance is assessed using three complementary metrics:
\begin{itemize}
    \item \textbf{Mean Squared Error (MSE)}, measuring overall prediction accuracy;
    \item \textbf{Mean Absolute Error (MAE)}, providing a scale-robust measure of deviation;
    \item \textbf{Dynamic Time Warping (DTW)}, capturing temporal alignment between predicted and ground-truth sequences and enabling evaluation under potential phase shifts.
\end{itemize}
While MSE and MAE are widely used in existing benchmarks, DTW offers an additional perspective on temporal similarity and is particularly relevant for non-stationary and regime-shifting time series. 

Noted that for the imputation task, metrics are evaluated over the entire signal instead of only the masked positions, thereby assessing both missing value estimation and overall reconstruction quality.

\subsubsection{Baselines and implementation}
The proposed FM-TSP is compared against a set of representative state-of-the-art models for time-series analysis, including Dlinear \cite{zeng_are_2023}, LightTS \cite{zhang_less_2022}, PatchTST \cite{nie_time_2023}, iTransformer \cite{liu_itransformer_2024}.  
All baseline results are obtained using carefully reimplemented versions under identical experimental conditions.  
For all methods, reported results correspond to the average performance over multiple runs to mitigate the effect of random initialization.

\subsubsection{Results and Discussion} 
\begin{figure}[h]
    \centering
    \includegraphics[width=3in]{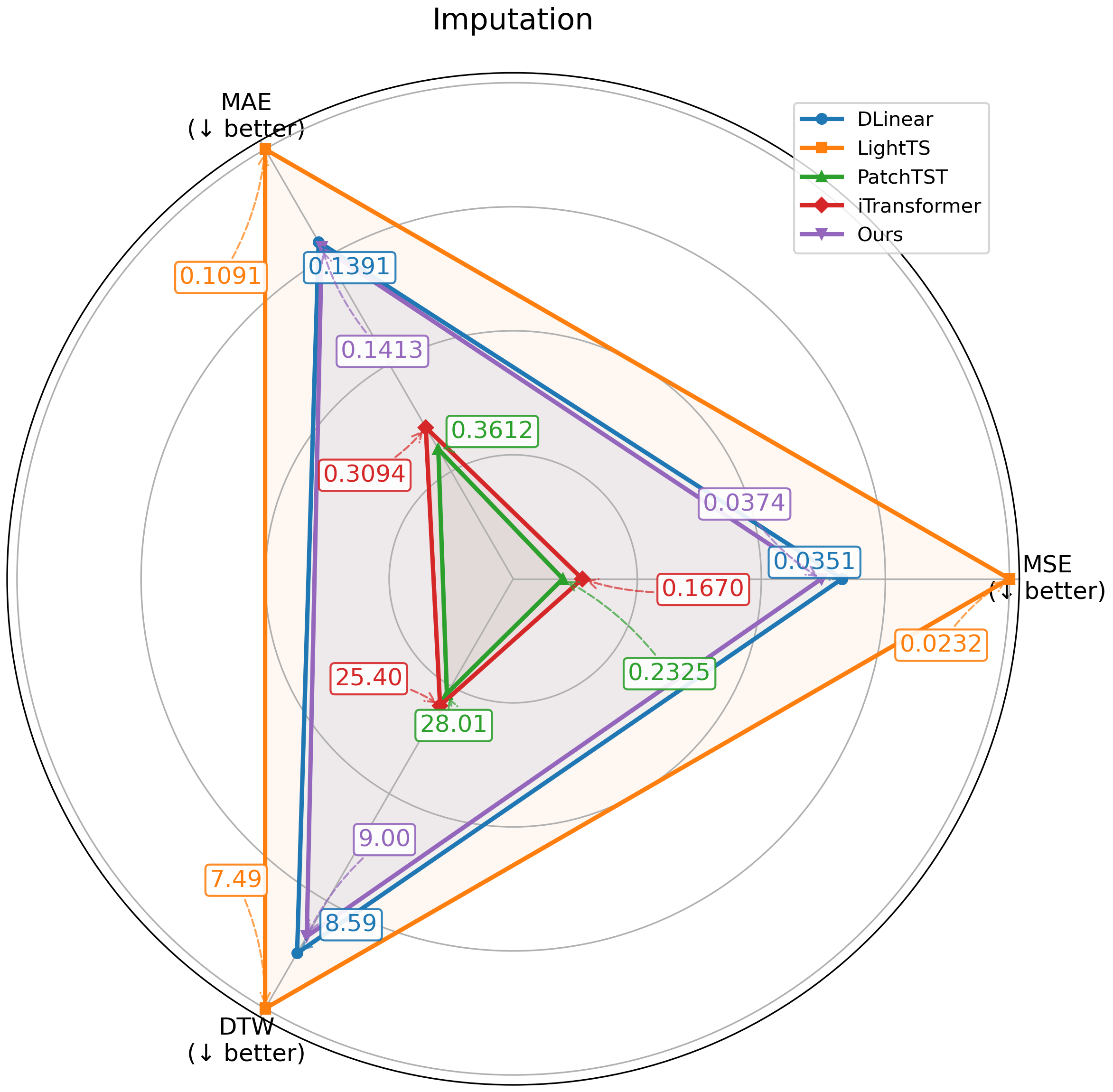}
    \caption{Performance comparison for imputation downstream task}
    \label{fig:imputation}
\end{figure}

\begin{figure}[h]
    \centering
    \includegraphics[width=3.4in]{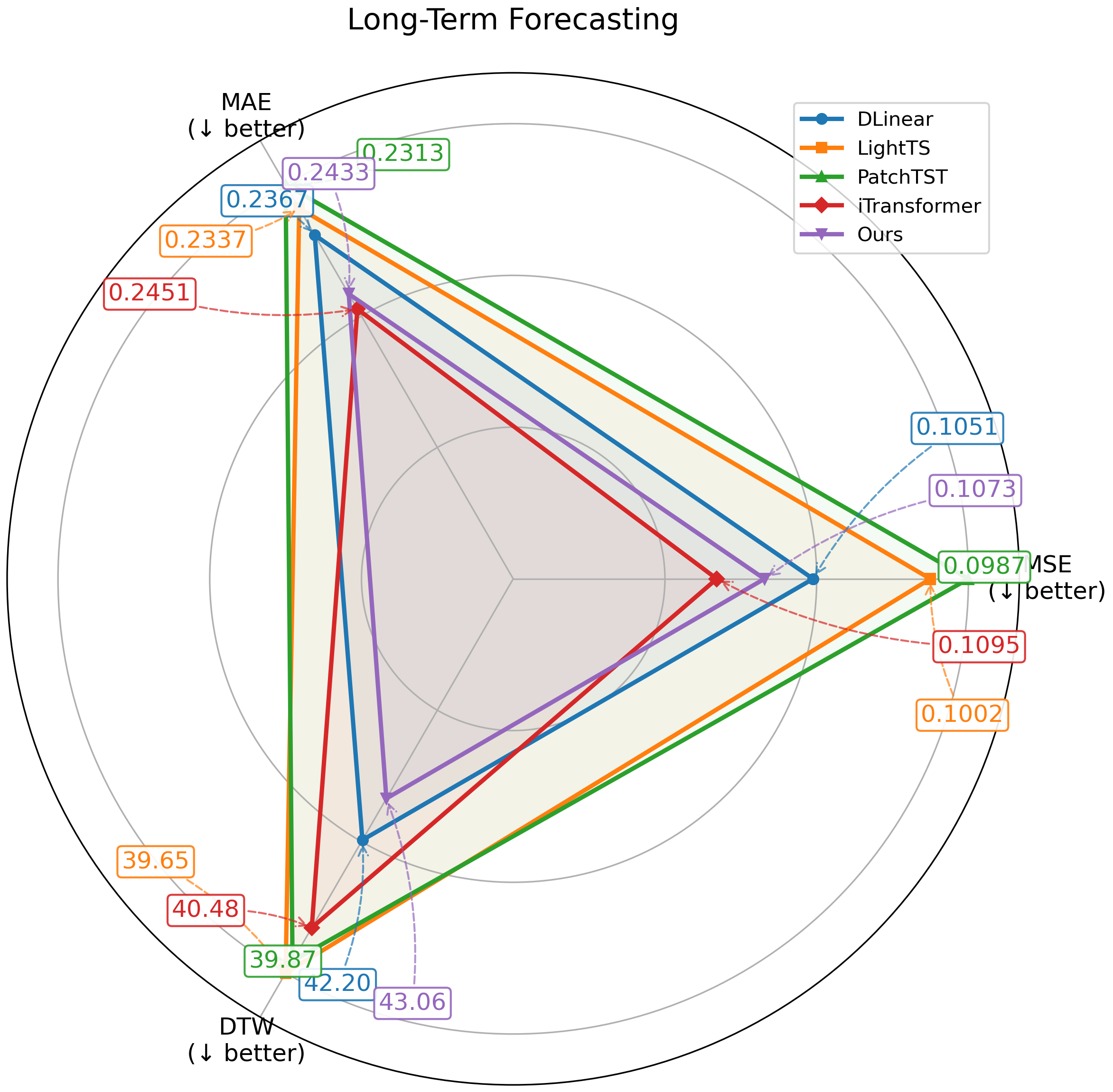}
    \caption{Performance comparison for long-term forecasting task}
    \label{fig:longterm}
\end{figure}

\begin{figure}[h]
    \centering
    \includegraphics[width=3.4in]{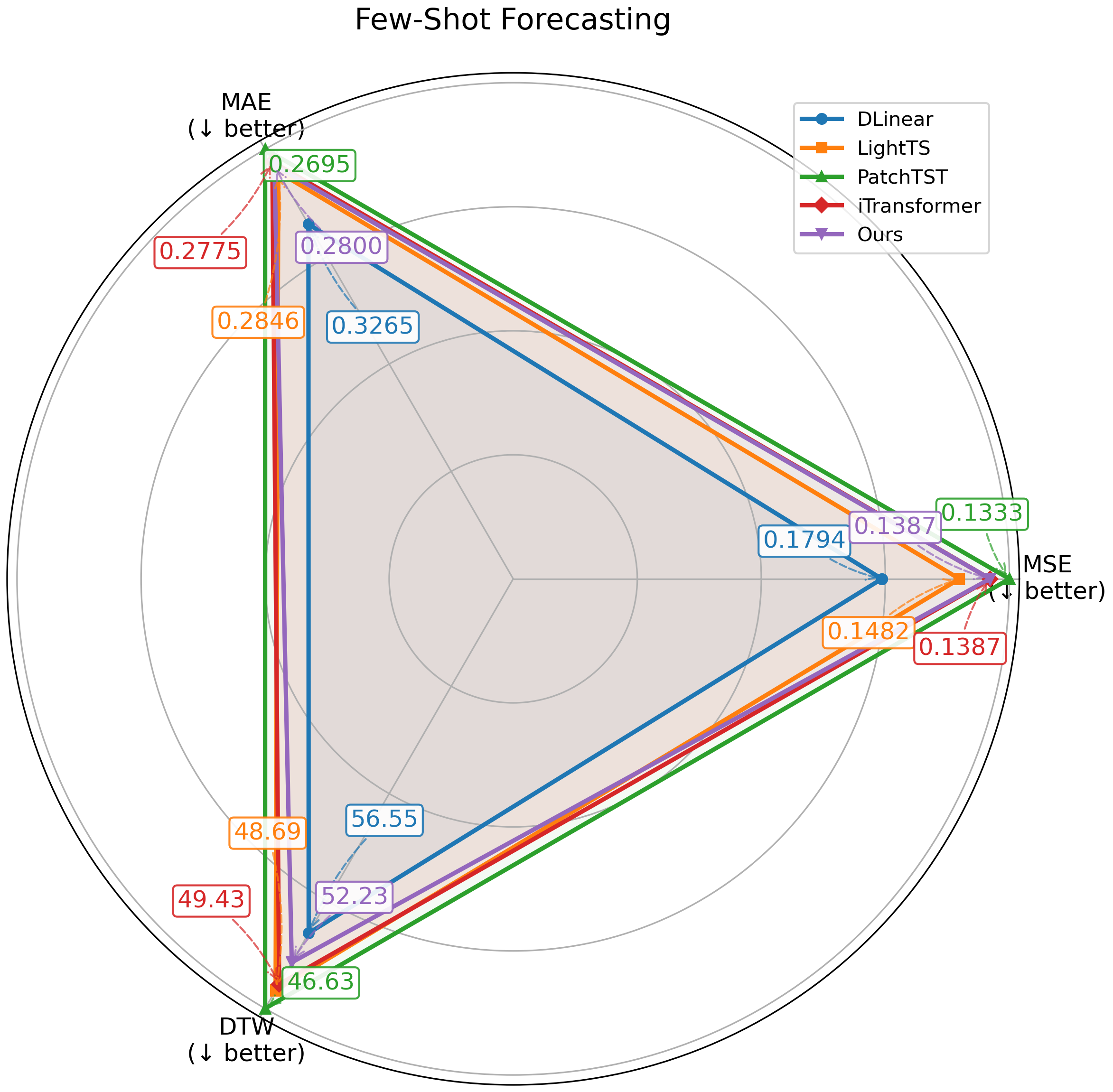}
    \caption{Performance comparison for few-shot learning forecasting task}
    \label{fig:fewshot}
\end{figure}

Figures~\ref{fig:imputation}–\ref{fig:fewshot} present the overall averages across all subsets (ETTh1, ETTh2, ETTm1, ETTm2) for each evaluation metric in imputation, long-term forecasting, and few-shot forecasting, respectively, and compare FM-TSP with several representative state-of-the-art baselines.

\paragraph{Imputation}
The summary of imputation performances in Figure~\ref{fig:imputation} shows that FM-TSP present a strong performance in this task. We outperform the PatchTST and iTransformer for all three metrics and have similar performance as Dlinear. This result is expected, as the imputation task is closely related to the encoder pretraining objective, namely the reconstruction task, implying that the encoders naturally possess certain advantages.
\paragraph{Long-term forecasting}
The performance of FM-TSP on this task (Figure~ \ref{fig:longterm}) is lower than that observed for the imputation task (Figure~\ref{fig:imputation}) or the few-shot forecasting task (Figure~\ref{fig:fewshot}). Nevertheless, FM-TSP remains in a competitive zone, with evaluation metrics close to those of DLinear and iTransformer. PatchTST and LightTS outperform FM-TSP on this task.
This result may be explained by the nature of the forecasting task, which requires capturing complex temporal relationships between past and future value features that may not be fully captured by the selected encoders. In fact, in this study, the encoders were not pretrained on the target ETT dataset. As a result, if none of the encoders can effectively extract key past–future dependencies, the performance in long-term forecasting may be affected.
However, this limitation can be addressed by extending the library of pretrained encoders with more suitable models. This highlights the flexibility of FM-TSP.

\paragraph{Few-shot forecasting}
Figure~\ref{fig:fewshot} presents the results of the few-shot forecasting experiments. This task is challenging for all models, as only $5\%$ and $10\%$ of the training data are available for fine-tuning. In this data-scarce setting, FM-TSP ranks among the top-performing models, alongside PatchTST, iTransformer, and LightTS, while DLinear performs comparatively worse.
This result demonstrates that the proposed framework effectively transfers knowledge from heterogeneous pretraining sources to downstream tasks, even in highly data-constrained scenarios.
\paragraph{Discussion}
Overall, these quantitative results demonstrate that FM-TSP performs robustly across reconstruction, prediction, and data-efficient adaptation scenarios.  Notably, these gains are achieved under a strict setting where the encoders are not pretrained on the target dataset, underscoring the framework’s ability to transfer generalized representations to downstream tasks without domain-specific pretraining.

Unlike conventional state-of-the-art methods that depend on end-to-end optimization and direct access to task-specific features from target data, FM-TSP uses frozen pretrained encoders and adapts only through downstream computation. Even with this limitation, it remains competitive with leading approaches, indicating that the transferred representations are sufficiently expressive to support diverse time-series inference objectives.

This behavior is especially well aligned with digital twin perception requirements, where a single deployed model must support multiple time-series inference tasks under varying data availability and operating conditions.

\subsection{Virtual sensor for rotor temperature - Industrial task case study}
\subsubsection{Stated Problem}
Accurate estimation of rotor temperature is essential for defining the operational limits of hydroelectric generators, as excessive temperatures accelerate degradation and restrict reactive power capability. Although direct temperature measurements provide the most reliable information, they are difficult to implement in practice due to rotor rotation and electrical safety constraints. To address this limitation, this study relies on a virtual sensor approach: a data-driven model calibrated using available measurements to estimate a target data indirectly. This virtual sensor is recently popularly developped in many industries for the machine health monitoring \cite{kahwati_building_2025}, \cite{gagnon_virtual_2022} or for further optimization processes \cite{gagnon_virtual_2025}. In this case study, by leveraging operational data recorded by the continuous monitoring system (CMS), the objective of temperature virtual sensor is to enable continuous assessment of rotor thermal behavior and identification of hot spots, even in non-instrumented or partially instrumented machines \cite{kahwati_building_2025}.

\subsubsection{Implementation}
We integrate the virtual sensor concept into our foundation model by modifying the output head. In particular, rather than predicting a single point estimate of the temperature, the output head predicts the parameters of a probability distribution, namely the mean and variance, using a Beta-loss likelihood function \cite{seitzer_pitfalls_2022}. Accordingly, the task-specific head is decomposed into two separate linear layers: one dedicated to estimating the mean and the other to estimating the variance. In contrast, other structures of foundation model remain unchanged.  FM-TSP processes multichannel time-series inputs composed of operational signals correlated with the target variable. Based on these inputs, the foundation model learns a latent representation of the generator’s operating state and is used to estimate the rotor temperature.

The virtual sensor model is trained on one turbine-generator unit (TGU) using a standard train–validation–test split. Its robustness is then assessed by evaluating the pretrained model on additional TGUs.

\subsubsection{Results and discussion}
\begin{figure}[h]
    \centering
    \includegraphics[width=5in]{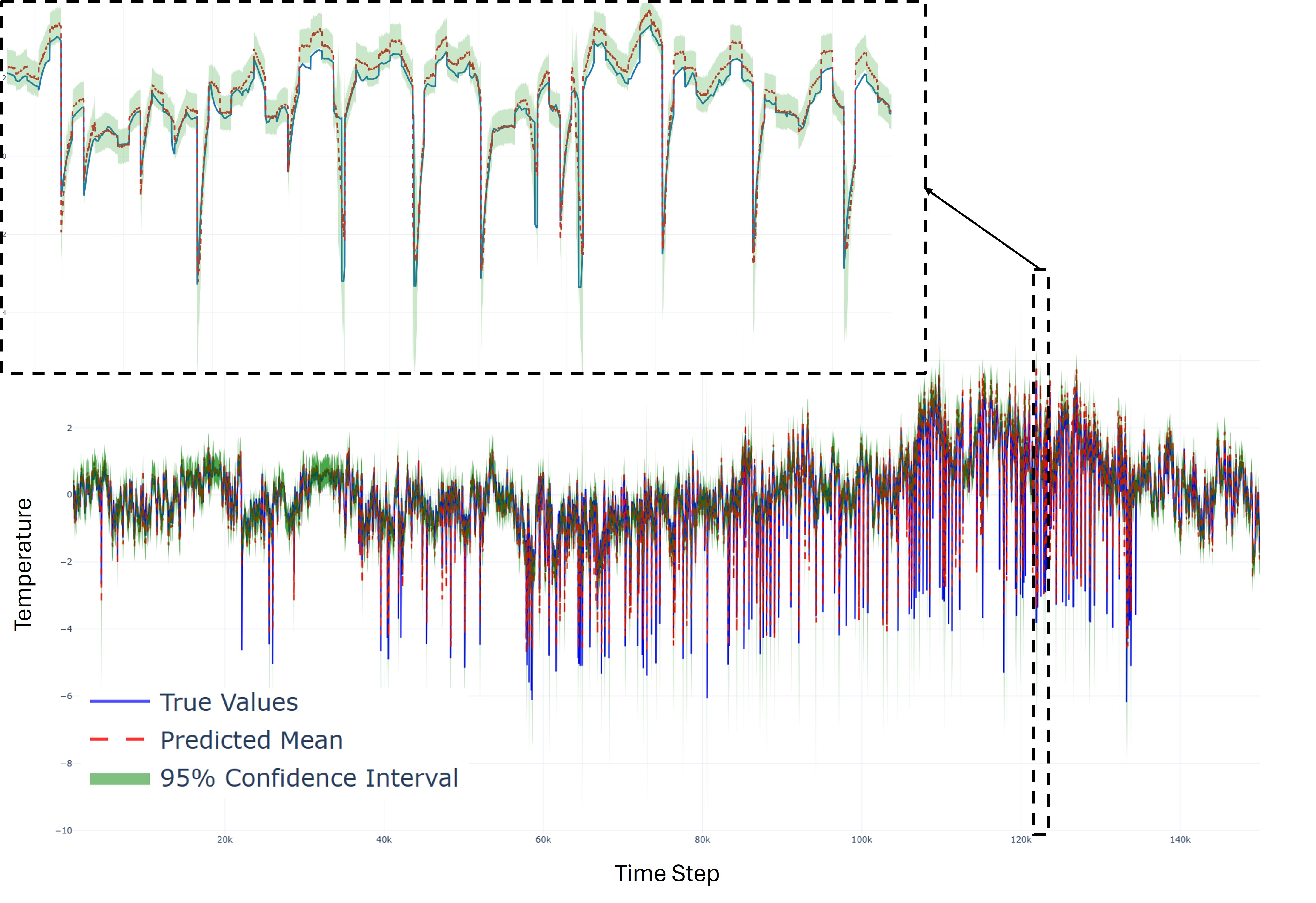}
    \caption{Temperature estimated by virtual sensor at training TGU. The zoomed-in section is taken from the split test set}
    \label{fig:vs_train_result}
\end{figure}

The prediction results presented in Figure~\ref{fig:vs_train_result} indicate successful model training. The predicted mean closely tracks the ground truth, while the $95\%$ confidence interval (95CI), derived from the predicted variance, effectively covers extreme values with low dispersion. A coverage level, that is calculated by Eq. \ref{eq_coverage}, of approximately $93.5\%$, which is close to the nominal $95\%$, shows that the uncertainty is well calibrated: 

\begin{equation}
\text{Coverage} = \frac{1}{N} \sum_{i=1}^N 
\mathbf{1}\left( \mathbf{T}_i \in [\hat{\mathbf{T}}_i^{lower}, \hat{\mathbf{T}}_i^{upper}] \right)
\label{eq_coverage}
\end{equation}
, where $\mathbf{T}$ is the temperature.

\begin{figure}[h]
    \centering
    \includegraphics[width=5in]{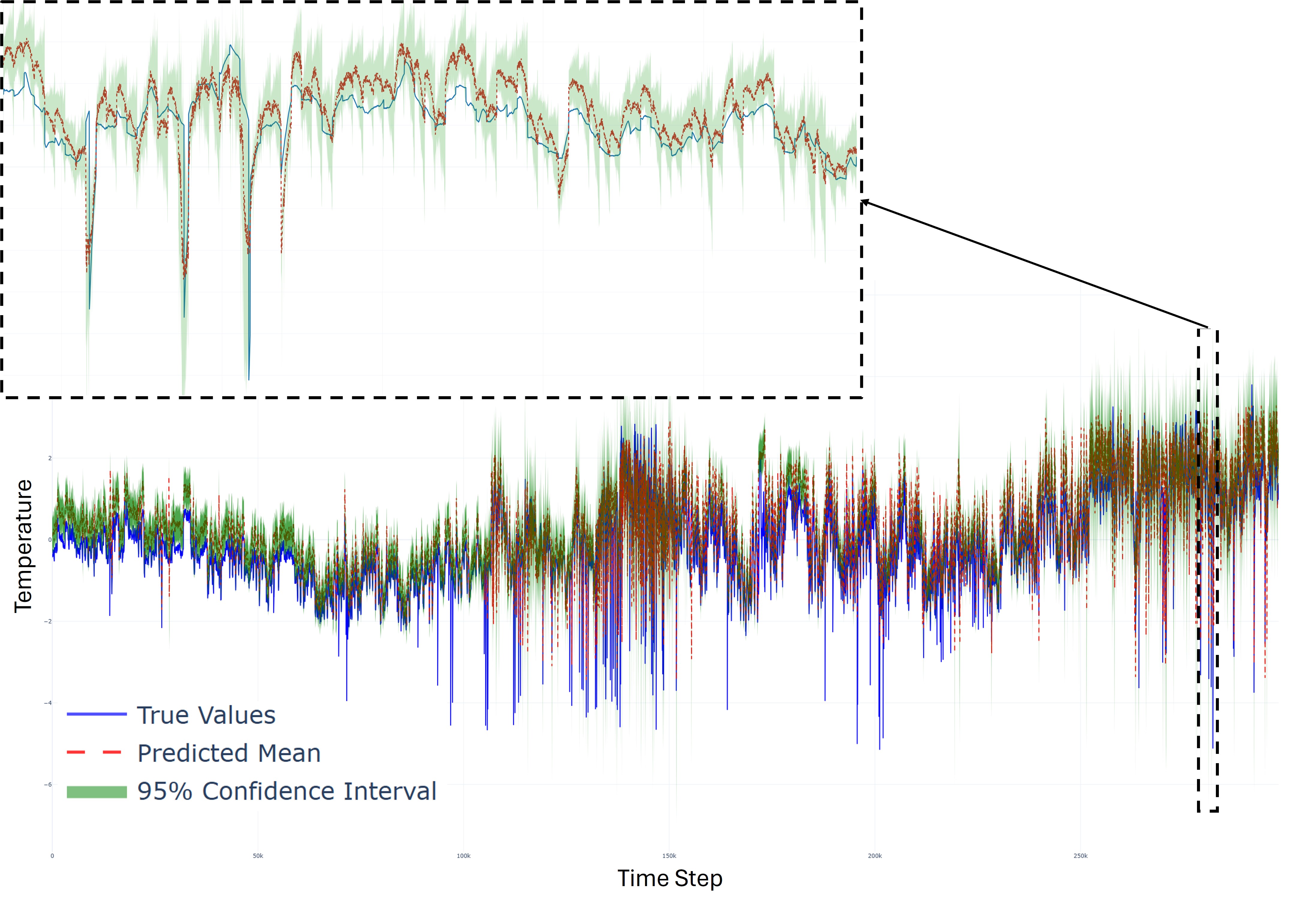}
    \caption{Temperature estimated by virtual sensor at test TGU}
    \label{fig:vs_test_result}
\end{figure}

Figure~\ref{fig:vs_test_result} presents the predictions on a different TGU using the pretrained model. The predicted mean continues to follow the overall trend of the ground truth, closely approximating its moving average. The 95CI indicates higher predicted variance. The lower coverage ($65\%$) is expected in this transfer learning task, which can be explained by the distribution shift between the two datasets. Indeed, two identical TGU designs may exhibit different behaviors, and different measurement systems can have varying sensitivities. As illustrated in Figure~\ref{fig:vs_test_result}, the temperature tends to be overestimated, particularly at extreme low peaks. However, in practice, extreme high temperatures are of greater importance for health monitoring, as they have a more significant impact on system reliability. Despite the wider confidence intervals, earlier detection of critical high-temperature events is generally more desirable in practice than delayed detection.

%%%%%%%%%%%%%%%%%%%%%%%%%%%%%%%%%%%%%%%%%%%%%%%%%%%%%%%%%
\section{Conclusion}\label{sec7}
This paper introduced a modular foundation model for time-series perception based on a collection of pretrained representation encoders. By leveraging self-supervised learning on heterogeneous datasets, the proposed framework learns transferable and task-agnostic representations that can be reused across multiple downstream tasks. The architecture explicitly decouples representation learning, encoder selection, and task adaptation, enabling scalable deployment under limited supervision.

A key contribution lies in the encoder-level gating mechanism, which dynamically selects relevant encoders based on data similarity and structural consistency. Combined with projection and Transformer-based aggregation, the proposed design effectively integrates heterogeneous representations. Ablation studies confirm that encoder selection and adaptive aggregation provide the largest performance gains, while representation alignment improves training stability.

Experimental results on the ETT benchmark demonstrate competitive performance across imputation, long-term forecasting, and few-shot learning tasks. A real-world industrial case study further highlights the practical relevance of the approach for digital twin perception and its ability to adapt to challenging settings involving multichannel time-series inputs, even when such signals are not encountered during the encoder pretraining phase.

Overall, the results suggest that scalable time-series perception can be achieved through modular and reusable representation learning rather than task-specific models. The proposed framework can also be naturally integrated into hybrid model–data PHM systems, combining data-driven representations with physics-based models to improve robustness and decision-making in industrial applications.

\section*{Author contributions}

\textbf{Quang Hung Pham}: Writing – original draft, Writing – review \& editing, Conceptualization, Methodology, Visualization, Software,  Investigation, Data curation. \textbf{Ryad Zemouri}: Writing – original draft, Writing – review \& editing, Conceptualization, Visualization, Methodology, Investigation. \textbf{Martin Gagnon}: Writing – review \& editing, Conceptualization, Methodology, Investigation. \textbf{Luc Vouligny}: Writing – review \& editing, Investigation. 

\section*{Acknowledgments}
The authors would like to thank Mehdi Lagnaoui (Université de Montréal) for his contribution to implementing and testing state-of-the-art methods, and Ghofril Kahwati (Hydro-Québec) for data-source preparation in the industrial case study.

The authors used AI tools such as Copilot to improve language editing. All research concept, development, validation, and final results were performed by the authors.

\bibliographystyle{unsrtnat}
\bibliography{references}  %%% Uncomment this line and comment out the ``thebibliography'' section below to use the external .bib file (using bibtex) .





% \end{thebibliography}

\end{document}